\documentclass[runningheads,oribibl]{llncs}
\newcommand{\papertitle}{
Exploiting Verified Neural Networks via Floating Point Numerical Error
}

\usepackage{mathtools}
\usepackage{amsmath}
\usepackage{amssymb}
\usepackage{multirow}
\usepackage{setspace}
\usepackage{threeparttable}
\usepackage{stfloats}
\usepackage{textcomp}
\usepackage[inline]{enumitem}
\usepackage{algorithm}
\usepackage{algpseudocode}
\usepackage{caption}
\usepackage{subcaption}
\usepackage{todonotes}
\usepackage{multirow}
\usepackage{bm}
\usepackage{bbm}
\usepackage{graphicx}
\usepackage[hidelinks, pdftex,
    pdfauthor={Kai Jia and Martin Rinard},
    pdftitle={\papertitle}]{hyperref}
\usepackage{url}
\usepackage{booktabs}       
\usepackage{amsfonts}       

\newcommand{\V}[1]{\bm{#1}}
\newcommand{\linf}{\ell_{\infty}}
\newcommand{\real}{\mathbb{R}}
\DeclareMathOperator{\argmax}{argmax}
\DeclareMathOperator{\argmin}{argmin}
\newcommand{\nncmd}[3]{\operatorname{#1}\left(#2;\,#3\right)}
\newcommand{\nncmdv}[3]{\nncmd{#1}{\V{#2}}{\V{#3}}}
\newcommand{\nn}[2]{\nncmdv{NN}{#1}{#2}}
\newcommand{\nncpuconv}[2]{\nncmdv{NN_{C,C}}{#1}{#2}}
\newcommand{\nncpumm}[2]{\nncmdv{NN_{C,M}}{#1}{#2}}
\newcommand{\nngpuconv}[2]{\nncmdv{NN_{G,C}}{#1}{#2}}
\newcommand{\nngpuconvwg}[2]{\nncmdv{NN_{G,CWG}}{#1}{#2}}
\newcommand{\nngpumm}[2]{\nncmdv{NN_{G,M}}{#1}{#2}}
\newcommand{\nnimpl}[2]{\nncmdv{NN_{impl}}{#1}{#2}}
\newcommand{\tauimpl}{\operatorname{\tau_{impl}}}
\newcommand{\cw}[2]{L_{\operatorname{CW}}\left(#1,\,#2\right)}
\newcommand{\cwv}[2]{\cw{\V{#1}}{#2}}
\newcommand{\xadv}{\V{x_{\operatorname{adv}}}}

\newcommand{\xseed}{\V{x_{\operatorname{seed}}}}
\newcommand{\adv}[1]{\operatorname{Adv}_{\epsilon}\left(\V{#1}\right)}
\newcommand{\floatdouble}[1]{\operatorname{float64}\left(#1\right)}

\newcommand{\condSet}[2]{\left\{#1 \;\middle|\; #2\right\}}
\newcommand{\norminf}[1]{\left\lVert#1\right\rVert_{\infty}}
\newcommand{\scinum}[2]{{#1}\mathrm{e}{#2}}
\newcommand{\defeq}{\vcentcolon=}

\usepackage[nameinlink,noabbrev]{cleveref}
\renewcommand{\cref}[1]{\Cref{#1}}
\newcommand{\eqnref}[1]{\hyperref[eqn:#1]{(\ref*{eqn:#1})}}

\newcommand{\vx}{\V{x}}
\newcommand{\vxz}{\V{x_0}}
\newcommand{\vxo}{\V{x_1}}
\newcommand{\vy}{\V{y}}

\newenvironment{enuminline}
{\begin{enumerate*}[label=(\roman*),itemjoin={{, }},itemjoin*={{, and }}]}
{\end{enumerate*}}

\newenvironment{enuminlinesemi}
{\begin{enumerate*}[label=(\roman*),itemjoin={{; }}]}{\end{enumerate*}}

\DeclareCaptionFormat{algor}{%
  \hrulefill\par\offinterlineskip\vskip1pt%
    \textbf{#1#2}#3\offinterlineskip\hrulefill}
\DeclareCaptionStyle{algori}{singlelinecheck=off,format=algor,labelsep=space}
\usepackage[numbers,sort&compress]{natbib}

\makeatletter
\renewcommand\@biblabel[1]{#1.}
\makeatother

\begin{document}
\title{\papertitle}
%
%
\author{Kai Jia\inst{1}\and
Martin Rinard\inst{1}}
%
%
\institute{MIT CSAIL, Cambridge MA 02139, USA \\
    \email{\{jiakai,rinard\}@mit.edu}
}

\maketitle              

\begin{abstract}
    Researchers have developed neural network verification algorithms motivated
    by the need to characterize the robustness of deep neural networks.  The
    verifiers aspire to answer whether a neural network guarantees certain
    properties with respect to all inputs in a space. However, many verifiers
    inaccurately model floating point arithmetic but do not thoroughly discuss
    the consequences.

    We show that the negligence of floating point error leads to  unsound
    verification that can be systematically exploited in practice. For a
    pretrained neural network, we present a method that efficiently searches
    inputs as witnesses for the incorrectness of robustness claims made by a
    complete verifier. We also present a method to construct neural network
    architectures and weights that induce wrong results of an incomplete
    verifier. Our results highlight that, to achieve practically reliable
    verification of neural networks, any verification system must accurately (or
    conservatively) model the effects of any floating point computations in the
    network inference or verification system.

\end{abstract}


\section{Introduction}

Deep neural networks (DNNs) have been successful at various tasks, including
image processing, language understanding, and robotic control \citep{
raghu2020aso}. However, they are vulnerable to adversarial inputs~\citep{
szegedy2014intriguing}, which are input pairs indistinguishable to human
perception that cause a DNN to give substantially different predictions.  This
situation has motivated the development of network verification algorithms that
claim to prove the robustness of a network~\citep{ bunel2020branch,
tjeng2018evaluating, salman2019convex}, specifically that the network produces
identical classifications for all inputs in a perturbation space around a given
input.

Verification algorithms typically reason about the behavior of the network
assuming real-valued arithmetic. In practice, however, the computation of both
the verifier and the neural network is performed on physical computers that use
floating point numbers and floating point arithmetic to approximate the
underlying real-valued computations. This use of floating point introduces
numerical error that can potentially invalidate the guarantees that the
verifiers claim to provide.  Moreover, the existence of multiple software and
hardware systems for DNN inference further complicates the situation because
different implementations exhibit different numerical error characteristics.
Unfortunately, prior neural network verification research rarely discusses
floating point (un)soundness issues (\cref{sec:background}).

This work considers two scenarios for a decision-making system relying on
verified properties of certain neural networks:
\begin{enuminlinesemi}
    \item The adversary can present arbitrary network inputs to the system
        while the network has been pretrained and fixed
    \item The adversary can present arbitrary inputs and also network weights
        and architectures to the system
\end{enuminlinesemi}.
We present an efficient search technique to find witnesses of the unsoundness of
complete verifiers under the first scenario. The second scenario enables
inducing wrong results more easily, as will be shown in
\cref{sec:exploit-incomplete}. Note that even though allowing arbitrary network
architectures and weights is a stronger adversary, it is still practical. For
example, one may deploy a verifier to decide whether to accept an untrusted
network based on its verified robustness, and an attacker might manipulate the
network so that its nonrobust behavior does not get noticed by the verifier.

Specifically, we train robust networks on the MNIST and CIFAR10 datasets. We
work with the \verb|MIPVerify| complete verifier~\citep{tjeng2018evaluating} and
several inference implementations included in the PyTorch
framework~\citep{NEURIPS2019_9015}. For each implementation, we construct image
pairs $(\vxz, \xadv)$ where $\vxz$ is a brightness-modified natural image, such
that the implementation classifies $\xadv$ differently from $\vxz$, $\xadv$
falls in a $\linf$-bounded perturbation space around $\vxz$, and the verifier
incorrectly claims that no such adversarial image $\xadv$ exists for $\vxz$
within the perturbation space.  Moreover, we show that if modifying network
architecture or weights is allowed, floating point error of an incomplete
verifier \verb|CROWN|~\citep{ zhang2018efficient} can also be exploited to
induce wrong results. Our method of constructing adversarial images is not
limited to our setting but is applicable to other verifiers that do not soundly
model floating point arithmetic.

We emphasize that any verifier that does not correctly or conservatively model
floating point arithmetic fails to provide any safety guarantee against
malicious network inputs and/or network architectures and weights.  Ad hoc
patches or parameter tuning can not fix this problem. Instead, verification
techniques should strive to provide soundness guarantees by correctly
incorporating floating point details in both the verifier and the deployed
neural network inference implementation. Another solution is to work with
quantized neural networks that eliminate floating point issues \citep{
jia2020efficient}.


\section{Background and related work} \label{sec:background}

\textbf{Training robust networks:} Researchers have developed various techniques
to train robust networks~\citep{ madry2018towards, mirman2018differentiable,
tramer2019adversarial, wong2020Fast}. \citet{ madry2018towards} formulates the
robust training problem as minimizing the worst loss within the input
perturbation and proposes training on data generated by the Projected Gradient
Descent (PGD) adversary. In this work, we consider robust networks trained with
the PGD adversary.

\noindent \textbf{Complete verification:} Complete verification (a.k.a.  exact
verification) methods either prove the property being verified or provide a
counterexample to disprove it. Complete verifiers have formulated the
verification problem as a Satisfiability Modulo Theories (SMT) problem~\citep{
scheibler2015towards, huang2017safety, katz2017reluplex, ehlers2017formal,
bunel2020branch} or a Mixed Integer Linear Programming (MILP) problem~\citep{
lomuscio2017approach, cheng2017maximum, fischetti2018deep, dutta2018output,
tjeng2018evaluating}. In principle, SMT solvers are able to model exact floating
point arithmetic~\citep{ rummer2010smt} or exact real arithmetic~\citep{
corzilius2012smt}. However, for efficiency reasons, deployed SMT solvers for
verifying neural networks all use inexact floating point arithmetic to reason
about the neural network inference. MILP solvers typically work directly with
floating point, do not attempt to model real arithmetic exactly, and therefore
suffer from numerical error. There have also been efforts on extending MILP
solvers to produce exact or conservative results \citep{ steffy2013valid,
neumaier2004safe}, but they exhibit limited performance and have not been
applied to neural network verification.

\noindent \textbf{Incomplete verification:} On the spectrum of the tradeoff
between completeness and scalability, incomplete methods (a.k.a. certification
methods) aspire to deliver more scalable verification by adopting
over-approximation while admitting the inability to either prove or disprove the
properties in certain cases. There is a large body of related research~\citep{
wong2017provable, weng2018towards, gehr2018ai2, zhang2018efficient,
raghunathan2018semidefinite, dvijotham2018training, mirman2018differentiable,
singh2019an}. \citet{salman2019convex} unifies most of the relaxation methods
under a common convex relaxation framework and suggests that there is an
inherent barrier to tight verification via layer-wise convex relaxation captured
by such a framework. We highlight that floating point error of implementations
that use a direct dot product formulation has been accounted for in some
certification frameworks~\citep{singh2018fast, singh2019an} by maintaining upper
and lower rounding bounds for sound floating point arithmetic~\citep{
mine2004relational}. Such frameworks should be extensible to model numerical
error in more sophisticated implementations like the Winograd convolution
\citep{ lavin2016fast}, but the effectiveness of this extension remains to be
studied.  However, most of the certification algorithms have not considered
floating point error and may be vulnerable to attacks that exploit this
deficiency.

\noindent \textbf{Floating point arithmetic:} Floating point is widely adopted
as an approximate representation of real numbers in digital computers. After
each calculation, the result is rounded to the nearest representable value,
which induces roundoff error. A large corpus of methods have been developed for
floating point analysis \citep{boldo2017computer, titolo2018abstract,
solovyev2018rigorous, das2020scalable}, but they have yet not been applied to
problems at the scale of neural network inference or verification involving
millions of operations. Concerns for floating point error in neural network
verifiers are well grounded. For example, the verifiers \verb|Reluplex|~\citep{
katz2017reluplex} and \verb|MIPVerify|~\citep{ tjeng2018evaluating} have been
observed to occasionally produce incorrect results on large scale benchmarks
\citep{wang2018formal, guidotti2020verification}. However, no prior work tries
to systematically invalidate neural network verification results via exploiting
floating point error.


\section{Problem definition}
\label{sec:adv-def}

We consider 2D image classification problems. Let $\vy=\nn{x}{W}$ denote the
classification confidence given by a neural network with weight parameters
$\V{W}$ for an input $\vx$, where $\vx \in \real_{[0,1]}^{m\times n \times c}$
is an image with $m$ rows and $n$ columns of pixels each containing $c$ color
channels represented by floating point values in the range $[0, 1]$, and $\vy
\in \real^k$ is a logits vector containing the classification scores for each of
the $k$ classes. The class with the highest score is the classification result
of the neural network.

For a logits vector $\vy$ and a target class number $t$, we define the
Carlini-Wagner (CW) loss~\citep{carlini2017towards} as the score of the target
class subtracted by the maximal score of the other classes:
\begin{align}
    \cwv{y}{t} = y_t - \max_{i\neq t}y_i
\end{align}
Note that $\vx$ is classified as an instance of class $t$ if and only if
$\cw{\nn{x}{W}}{t} > 0$, assuming no equal scores of two classes.

Adversarial robustness of a neural network is defined for an input $\vxz$ and a
perturbation bound $\epsilon$, such that the classification result is stable
within allowed perturbations:
\begin{align}
    \arraycolsep=.5ex
    \begin{array}{rl}
        \forall \vx\in\adv{x_0}:\: & L(\vx) > 0 \\
        \text{where } L(\vx) &= \cw{\nn{x}{W}}{t_0} \\
        t_0 &= \argmax \nn{x_0}{W}
    \end{array}
    \label{eqn:robustness-def}
\end{align}

In this work we consider $\linf$-norm bounded perturbations:
\begin{align}
    \adv{x_0} = \condSet{\vx}{\norminf{\vx-\vxz}\leq \epsilon \;\land\;
    \min \vx \ge 0 \;\land\; \max \vx \le 1}
\end{align}

We use the \verb|MIPVerify| \citep{tjeng2018evaluating} complete verifier to
demonstrate our attack method. \verb|MIPVerify| formulates
\eqnref{robustness-def} as an MILP instance $L^*=\min_{\vx\in\adv{x_0}} L(\vx)$
that is solved by the commercial solver Gurobi \citep{gurobi}. The network is
robust if $L^*>0$.  Otherwise, the minimizer $\vx^*$ encodes an adversarial
image.

Due to the inevitable presence of numerical error in both the network inference
system and the verifier, the exact specification of $\nn{\cdot}{W}$ (i.e., a
bit-level accurate description of the underlying computation) is not clearly
defined in \eqnref{robustness-def}. We consider the following implementations
included in the PyTorch framework to serve as our candidate definitions of the
convolutional layers in $\nn{\cdot}{W}$, while nonconvolutional layers use the
default PyTorch implementation:
\begin{itemize}
    \item $\nncpumm{\cdot}{W}$: A matrix-multiplication-based implementation on
        x86/64 CPUs. The convolution kernel is copied into a matrix that
        describes the dot product to be applied on the flattened input for each
        output value.
    \item $\nncpuconv{\cdot}{W}$: The default convolution implementation on
        x86/64 CPUs.
    \item $\nngpumm{\cdot}{W}$: A matrix-multiplication-based implementation on
        NVIDIA GPUs.
    \item $\nngpuconv{\cdot}{W}$: A convolution implementation using the
        \verb|IMPLICIT_GEMM| algorithm from the cuDNN library~\citep{
        chetlur2014cudnn} on NVIDIA GPUs.
    \item $\nngpuconvwg{\cdot}{W}$: A convolution implementation using the
        \verb|WINOGRAD_NONFUSED| algorithm from the cuDNN library~\citep{
        chetlur2014cudnn} on NVIDIA GPUs. It is based on the Winograd
        convolution algorithm~\citep{lavin2016fast}, which runs faster but has
        higher numerical error compared to others.
\end{itemize}

For a given implementation $\nnimpl{\cdot}{W}$, our method finds pairs of
$(\vxz,\, \xadv)$ represented as single precision floating point numbers such
that
\begin{enumerate}
    \item $\vxz$ and $\xadv$ are in the dynamic range of images: \\
        \centerline{$\min\vxz \ge 0$, $\min \xadv \ge 0$, $\max \vxz \le 1$,
        and $\max \xadv \le 1$}
    \item $\xadv$ falls in the perturbation space of $\vxz$: $\norminf{
        \xadv-\vxz} \le \epsilon$
    \item The verifier claims that the robustness specification
        \eqnref{robustness-def} holds for $\vxz$
    \item The implementation falsifies the claim of the verifier: \\
        \centerline{$\cw{\nnimpl{\xadv}{W}}{t_0} < 0$}
\end{enumerate}

Note that the first two conditions are accurately defined for any implementation
compliant with the IEEE-754 standard \citep{ieee754}, because the computation
only involves element-wise subtraction and max-reduction that incur no
accumulated error. The \verb|Gurobi| solver used by \verb|MIPVerify| operates
with double precision internally. Therefore, to ensure that our adversarial
examples satisfy the constraints considered by the solver, we also require that
the first two conditions hold for $\xadv'=\floatdouble{\xadv}$ and
$\vxz'=\floatdouble{\vxz}$ that are double precision representations of $\xadv$
and $\vxz$.


\section{Exploiting a complete verifier}
\label{sec:method-complete}

We present two observations crucial to the exploitation to be described later.

\begin{figure}[htb]
    \centering
    \includegraphics[width=\linewidth]{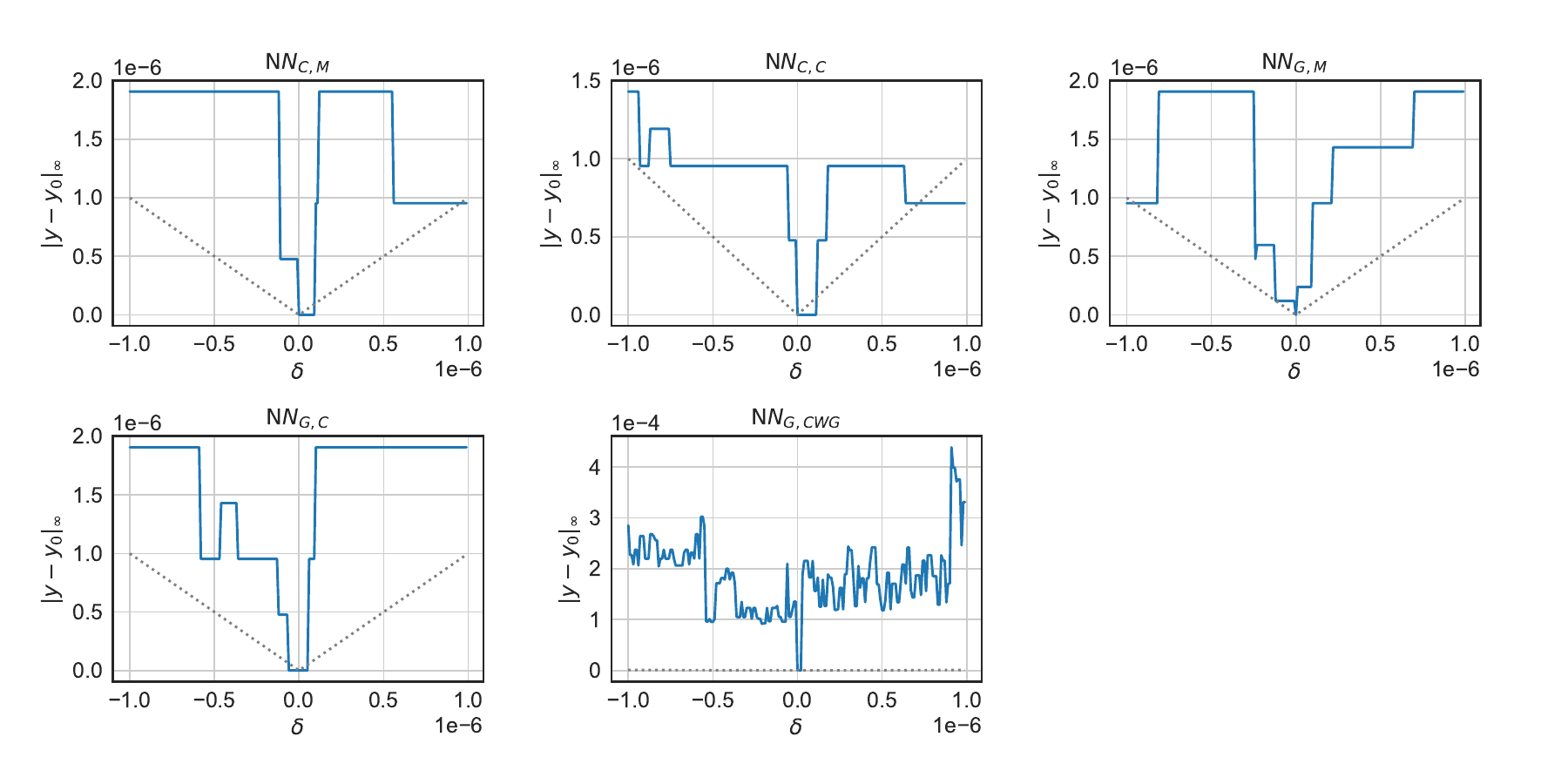}
    \caption{Change of logits vector due to small single-element input
        perturbations for different implementations. The dashed lines are $y =
        |\delta|$.
        \label{fig:e2e-err}
    }
\end{figure}

\noindent
\textbf{Observation 1:} Tiny perturbations on the network input result in random
output perturbations. We select an image $\vx$ for which the verifier claims
that the network makes robust predictions. We plot $\norminf{\nncmd{NN}
{\vx+\delta}{\V{W}} - \nn{x}{W}}$ against $-10^{-6} \le\delta\le10^{-6}$, where the
addition of $\vx+\delta$ is only applied on the single input element that has
the largest gradient magnitude. As shown in \cref{fig:e2e-err}, the change of
the output is highly nonlinear with respect to the change of the input, and a
small perturbation could result in a large fluctuation. Note that the output
fluctuation is caused by accumulated floating point error instead of
nonlinearities in the network because pre-activation values of all the ReLU
units have the same signs for both $\vx$ and $\vx+\delta$.

\begin{figure}[htb]
    \centering
    \includegraphics[width=.8\linewidth]{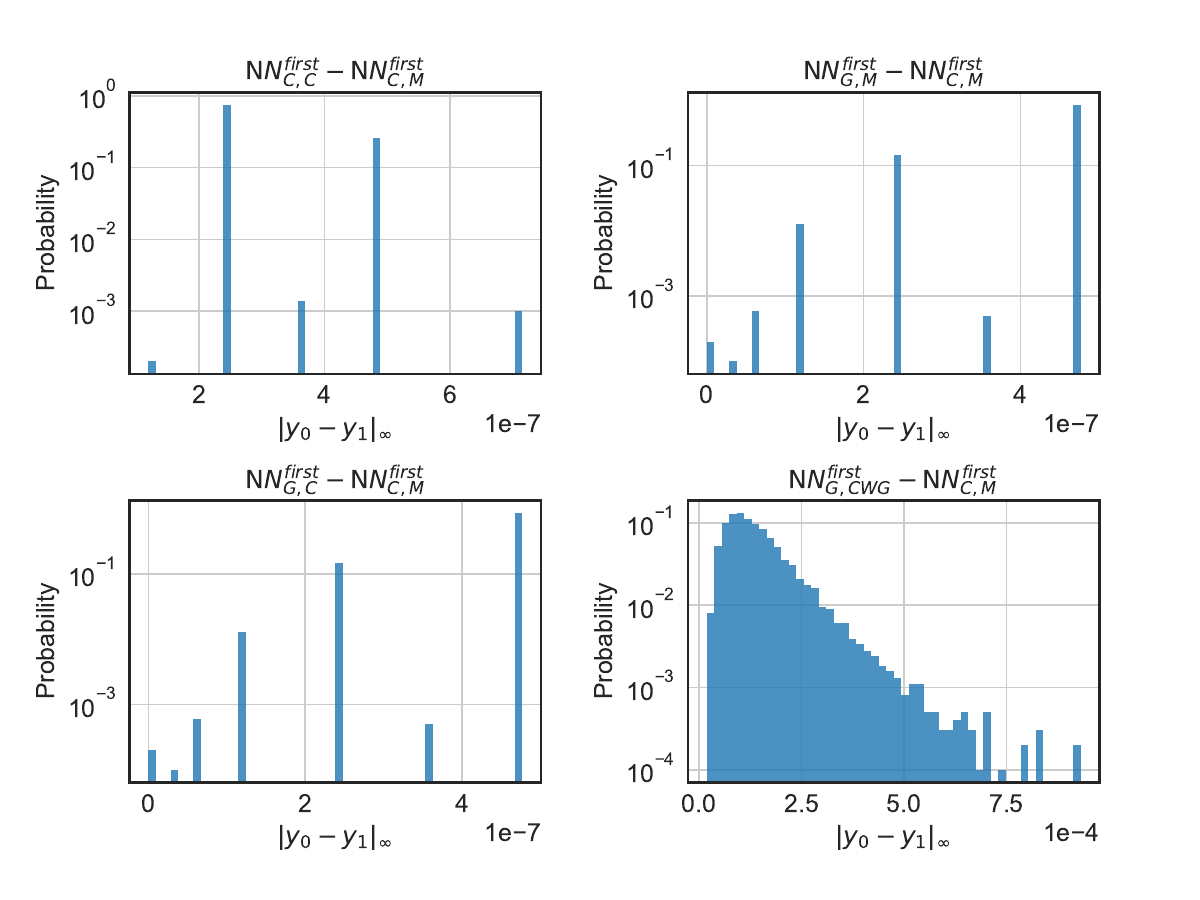}
    \caption{Distribution of difference relative to $\operatorname{NN_{C,M}}$ of
        first layer evaluated on MNIST test images.
        \label{fig:rela-err}
    }
\end{figure}

\noindent
\textbf{Observation 2:} Different neural network implementations exhibit
different floating point error characteristics. We evaluate the implementations
on the whole MNIST test set and compare the outputs of the first layer (i.e.,
with only one linear transformation applied to the input) against that of
$\operatorname{NN_{C,M}}$. \cref{fig:rela-err} presents the histogram which
shows that different implementations usually manifest different error behavior.

\begin{figure}[t]
    \centering
    \includegraphics[width=.8\linewidth]{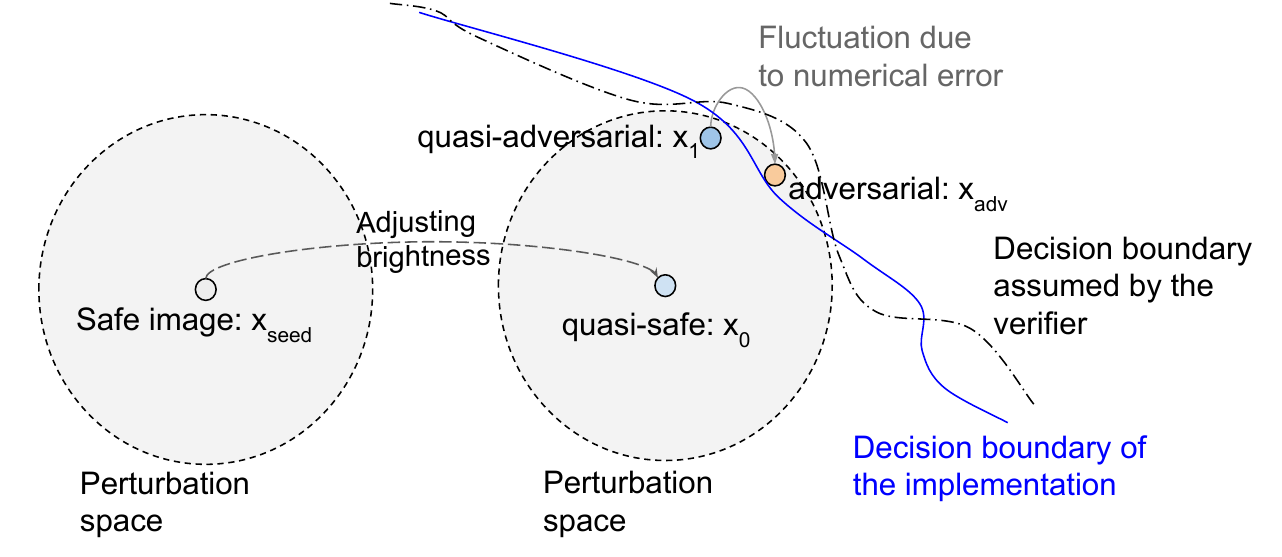}
    \caption{Illustration of our method. Since the verifier does not model the
        floating point arithmetic details of the implementation, their decision
        boundaries for the classification problem diverge, which allows us to
        find adversarial inputs that cross the boundary via numerical error
        fluctuations. Note that the verifier usually does not comply with a well
        defined specification of $\nn{\cdot}{W}$, and therefore it does not
        define a decision boundary.  The dashed boundary in the diagram is just
        for illustrative purposes.}
    \label{fig:method}
\end{figure}

\noindent
\textbf{Method Overview:} Given a network and weights $\nn{\cdot}{W}$, we search
for image pairs $(\vxz, \vxo)$ such that the network is verifiably robust with
respect to $\vxz$, while $\vxo \in \adv{\vxz}$ and $\cw{\nn{x_1}{W}}{t_0}$ is
less than the numerical fluctuation introduced by tiny input perturbations. We
call $\vxz$ a \emph{quasi-safe image} and $\vxo$ the corresponding
\emph{quasi-adversarial image}.  Observation 1 suggests that an adversarial
image might be obtained by randomly disturbing the quasi-adversarial image in
the perturbation space. Observation 2 suggests that each implementation has its
own adversarial images and needs to be handled separately. We search for the
quasi-safe image by modifying the brightness of a natural image while querying a
complete verifier whether it is near the boundary of robust predictions.
\cref{fig:method} illustrates this process.

Before explaining the details of our method, we first present the following
proposition that formally establishes the existence of quasi-safe and
quasi-adversarial images for continuous neural networks:
\begin{proposition}
    Let $E > 0$ be an arbitrarily small positive number. If a continuous neural
    network $\nn{\cdot}{W}$ can produce a robust classification for some input
    belonging to class $t$, and it does not constantly classify all inputs as
    class $t$, then there exists an input $\vxz$ such that
    \begin{align*}
        0 < \min_{\vx\in\adv{\vxz}} \cw{\nn{x}{W}}{t} < E
    \end{align*}
    Let $\vxo = \argmin_{\vx\in\adv{\vxz}} \cw{\nn{x}{W}}{t}$ be the
    minimizer of the above function. We call $\vxz$ a quasi-safe image and
    $\vxo$ a quasi-adversarial image.
\end{proposition}
\begin{proof}
    Let $f(\vx) \defeq \min_{\vx'\in\adv{x}} \cw{\nn{x'}{W}}{t}$. Since $f(\cdot)
    $ is composed of continuous functions, $f(\cdot)$ is continuous.  Suppose
    $\nn{\cdot}{W}$ is robust with respect to $\V{x_+}$ that belongs to class
    $t$. Let $\V{x_-}$ be be any input such that $\cw{\nn{x_-}{W}}{t} < 0$,
    which exists because $\nn{\cdot}{W}$ does not constantly classify all inputs
    as class $t$. We have $f(\V{x_+})>0$ and $f(\V{x_-}) < 0$. The
    Poincar\'e-Miranda theorem asserts the existence of $\vxz$ such that $0 <
    f(\vxz) < E$.
\end{proof}

Given a particular implementation $\nnimpl{\cdot}{W}$ and a natural image
$\xseed$ which the network robustly classifies as class $t_0$ according to the
verifier, we construct an adversarial input pair $(\vxz,\, \xadv)$ that meets
the constraints described in \cref{sec:adv-def} in three steps:

\noindent
\textbf{Step 1:} We search for a coefficient $\alpha\in[0,1]$ such that
$\vxz=\alpha \xseed$ serves as the quasi-safe image. Specifically, we require
the verifier to claim that the network is robust for $\alpha\xseed$ but not so
for $\alpha'\xseed$ with $0 < (\delta=\alpha-\alpha') < \epsilon_r$, where
$\epsilon_r$ should be small enough to allow quasi-adversarial images
sufficiently close to the boundary. We set $\epsilon_r=10^{-7}$. We use binary
search to minimize $\delta$ starting from $\alpha'\gets0$, $\alpha\gets1$. We
found that the MILP solver often becomes extremely slow when $\delta$ is small,
so we start with binary search and switch to grid search by dividing the best
known $\delta$ to 16 intervals if the solver exceeds a time limit.

\noindent
\textbf{Step 2:} \newcommand{\ltau}[2]{L(#1,\,#2)}
We search for the quasi-adversarial image $\vxo$ corresponding to $\vxz$. We
define a loss function with a tolerance of $\tau$ as $\ltau{\vx}{\tau} \defeq
\cw{\nn{x}{W}}{t_0}-\tau$, which can be incorporated in any verifier by
modifying the bias of the Softmax layer. We aim to find $\tau_0$ and $\tau_1$,
where $\tau_0$ is the minimal confidence of all images in the perturbation space
of $\vxz$, and $\tau_1$ is slightly larger than $\tau_0$ with $\vxo$ being the
corresponding adversarial image. Formally:
\begin{align*}
    \left\{\begin{array}{l}
            \forall \vx\in\adv{x_0}:\: \ltau{\vx}{\tau_0} > 0 \\
            \vxo \in \adv{x_0} \\
            \ltau{\vxo}{\tau_1} < 0 \\
            \tau_1 - \tau_0 < 10^{-7}
    \end{array}\right.
\end{align*}
Note that $\vxo$ is produced by the complete verifier as proof of nonrobustness
given the tolerance $\tau_1$. The above values are found via binary search
with initialization $\tau_0 \gets 0$ and $\tau_1 \gets \cw{\nn{x_0} {W}}{t_0}$.
In addition, we accelerate the binary search if the verifier can compute the
\emph{worst} objective defined as: \begin{align} \tau_w = \min_{\vx\in\adv{x_0}}
    \cw{\nn{x}{W}}{t_0}
\end{align}
In this case, we initialize $\tau_0 \gets \tau_w - \delta_s$ and $\tau_1 \gets
\tau_w + \delta_s$. We empirically set $\delta_s = 3\times10^{-6}$ to
incorporate the numerical error in the verifier so that $\ltau{\vxz}{
\tau_w-\delta_s} > 0$ and $\ltau{\vxz}{\tau_w+\delta_s} < 0$. The binary search
is aborted if the solver times out.

\noindent
\textbf{Step 3:} We minimize $\cw{\nn{x_1}{W}}{t_0}$ with hill climbing via
applying small random perturbations on the quasi-adversarial image $\vxo$ while
projecting back to $\adv{x_0}$ to find an adversarial example. The perturbations
are applied on patches of $\vxo$ as described in \cref{alg:rnd-pert-adv}.

\begin{center}
    \captionof{algorithm}{
        Searching adversarial examples via hill climbing
        \label{alg:rnd-pert-adv}
    }
\begin{algorithmic}
    \footnotesize
    \newcommand{\stride}{\operatorname{stride}}
    \newcommand{\offset}{\operatorname{offset}}
    \State {\bfseries Input:} quasi-safe image $\vxz$
    \State {\bfseries Input:} target class number $t$
    \State {\bfseries Input:} quasi-adversarial image $\vxo$
    \State {\bfseries Input:} input perturbation bound $\epsilon$
    \State {\bfseries Input:} a neural network inference implementation
        $\nnimpl{\cdot}{W}$
    \State {\bfseries Input:} number of iterations $N$ (default value $1000$)
    \State {\bfseries Input:} perturbation scale $u$ (default value
        $\scinum{2}{-7}$)
    \State {\bfseries Output:} an adversarial image $\xadv$ or FAILED
    \vspace{.4em}

    \For{Index $i$ of $\vxz$}
        \Comment {Find the bounds $\V{x_l}$ and $\V{x_u}$ for allowed
            perturbations}
        \State $x_l[i] \gets \max(\operatorname{nextafter}
            (x_0[i]-\epsilon,\,0),\, 0)$
        \State $x_u[i] \gets \min(\operatorname{nextafter}
            (x_0[i]+\epsilon,\,1),\, 1)$

        \While {${x_0[i] - x_l[i]} > \epsilon$ {\bf or}
            ${\floatdouble{x_0[i]} - \floatdouble{x_l[i]}} > \epsilon$}
            \State $x_l[i] \gets \operatorname{nextafter}(x_l[i], 1)$
        \EndWhile
        \While {${x_u[i] - x_0[i]} > \epsilon$ {\bf or}
            ${\floatdouble{x_u[i]} - \floatdouble{x_0[i]}} > \epsilon$}
            \State $x_u[i] \gets \operatorname{nextafter}(x_u[i], 0)$
        \EndWhile
    \EndFor
    \vspace{.4em}

    \Comment{
        We select the offset and stride based on the implementation to ensure
        that perturbed tiles contribute independently to the output.
        The Winograd algorithm in cuDNN produces $9\times9$ output
        tiles for $13\times13$ input tiles and $5\times5$ kernels.
    }
    \If{$\nnimpl{\cdot}{W}$ is $\nngpuconvwg{\cdot}{W}$}
        $(\offset,\,\stride) \gets (4, 9)$
    \Else
        \hspace{1em} $(\offset,\,\stride) \gets (0, 4)$
    \EndIf
    \vspace{.4em}

    \For{$i \gets 1$ {\bf to} $N$}
    \For{$(h,w) \gets (0,\,0)$ {\bf to}
            $(\operatorname{height}(\vxo),\,\operatorname{width}(\vxo))$
            {\bf step}
            $(\stride,\, \stride)$
        }
            \State $\delta \gets \operatorname{uniform}(-u,\,u,\, (
                \stride - \offset ,\, \stride - \offset))$
            \State $\vxo' \gets \vxo[:]$
            \State $\vxo'[h+\offset:h+\stride,\,w+\offset:w+\stride]
                \mathrel{+}\mathrel{\mkern-1mu}= \delta$
            \State $\vxo' \gets \max(\min(\vxo',\, \V{x_u}),\,\V{x_l})$

            \If{$\cw{\nnimpl{\vxo'}{W}}{t} < \cw{\nnimpl{\vxo}{W}}{t}$}
                \State $\vxo \gets \vxo'$
            \EndIf
        \EndFor
    \EndFor
    \If{$\cw{\nnimpl{\vxo}{W}}{t} < 0$}
        \Return{$\xadv \gets \vxo$}
    \Else
        \hspace{1em} \Return{FAILED}
    \EndIf
\end{algorithmic}
\hrulefill
\end{center}

\newcommand{\nrModelMnist}{18}
\newcommand{\nrModelCifar}{26}
\newcommand{\nrSuccTotal}{82}
\newcommand{\maxSuccLoss}{3.2\times10^{ -4 }}
\newcommand{\minFailLoss}{8.3\times10^{ -7 }}
\newcommand{\maxLoss}{3.7}
\newcommand{\nrSuccBelowFail}{35}
\newcommand{\nrSuccBelowZero}{18}

\noindent
\textbf{Experiments:} We conduct our experiments on a workstation with an NVIDIA
Titan RTX GPU and an AMD Ryzen Threadripper 2970WX CPU. We train the small
architecture from \citet{xiao2018training} with the PGD adversary and the RS
Loss on MNIST and CIFAR10 datasets. The network has two convolutional layers
with $4\times4$ filters, $2\times2$ stride, and 16 and 32 output channels,
respectively, and two fully connected layers with 100 and 10 output neurons. The
trained networks achieve 94.63\% and 44.73\% provable robustness with
perturbations of $\linf$ bounded by $0.1$ and $2/255$ on the two datasets,
respectively, similar to the results reported in~\citet{xiao2018training}. Our
code is available at \url{https://github.com/jia-kai/realadv}.

\begin{figure}[tb]
    \centering
    \includegraphics[clip,
        trim=0.25in 0.6in 0.25in 0.1in,width=\linewidth]{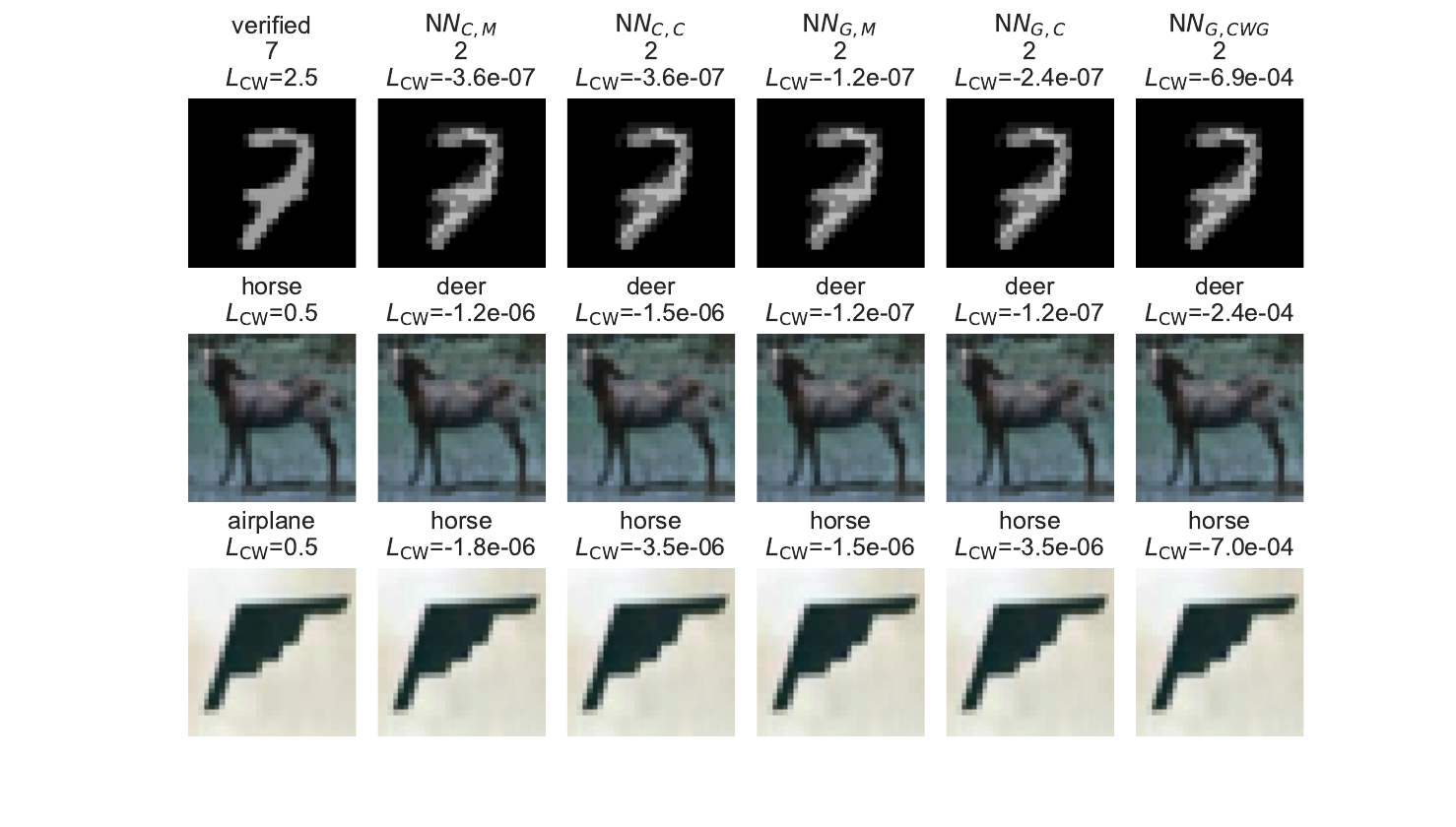}
    \caption{The quasi-safe images with respect to which all implementations are
        successfully exploited, and the corresponding adversarial images.}
    \label{fig:advimg}
\end{figure}

Although our method only needs $O(-\log \epsilon)$ invocations of the verifier
where $\epsilon$ is the threshold in the binary search, the verifier still takes
most of the time and is too slow for a large benchmark. Therefore, for each
dataset, we test our method on $32$ images randomly sampled from the verifiably
robustly classified test images. All the implementations that we have considered
are successfully exploited. Specifically, our benchmark contains
$32\times2\times5=320$ cases, while adversarial examples are found for
$\nrSuccTotal$ of them. The failed cases correspond to large $\tau_1$ values in
Step 2 due to verifier timeouts or the discrepancy of floating point arithmetic
between the verifier and the implementations. Let $\tauimpl \defeq \cw{
\nnimpl{\vxo}{W}}{t}$ denote the loss of an quasi-adversarial input on a
particular implementation. \cref{alg:rnd-pert-adv} succeeds on all cases with
$\tauimpl < \minFailLoss$ ($\nrSuccBelowFail$ such cases in total), while
$\nrSuccBelowZero$~among them have $\tauimpl < 0$ due to floating point
discrepancy (i.e., the quasi-adversarial input is already an adversarial input
for this implementation). The most challenging case (i.e., with largest
$\tauimpl$) on which \cref{alg:rnd-pert-adv} succeeds has
$\tauimpl=\maxSuccLoss$. The largest value of $\tauimpl$ is $\maxLoss$.
\cref{tab:adv-cnt} presents the detailed numbers for each implementation.
\cref{fig:advimg} shows the quasi-safe images on which our exploitation method
succeeds for all implementations and the corresponding adversarial images.

\begin{table}[tb]
    \caption{Number of adversarial examples successfully found for different
        neural network inference implementations
        \label{tab:adv-cnt}
    }
    \vspace{.5em}
    \centering
    \begin{tabular}{lrrrrr}
\toprule
{} & $\operatorname{NN_{C,M}}$ & $\operatorname{NN_{C,C}}$ & $\operatorname{NN_{G,M}}$ & $\operatorname{NN_{G,C}}$ & $\operatorname{NN_{G,CWG}}$ \\
\midrule
MNIST   &                         2 &                         3 &                         1 &                         3 &                           7 \\
CIFAR10 &                        16 &                        12 &                         7 &                         6 &                          25 \\
\bottomrule
\end{tabular}

    \vspace{1em}
\end{table}

\section{Exploiting an incomplete verifier}
\label{sec:exploit-incomplete}

The relaxation adopted in certification methods renders them incomplete but also
makes their verification claims more robust to floating point error compared to
complete verifiers. In particular, we evaluate the \verb|CROWN|
framework~\citep{ zhang2018efficient} on our randomly selected MNIST test images
and the corresponding quasi-safe images from \cref{sec:method-complete}.
\verb|CROWN| is able to verify the robustness of the network on $29$ out of the
$32$ original test images, but it is unable to prove the robustness for any of
the quasi-safe images. Note that \verb|MIPVerify| claims that the network is
robust with respect to all the original test images and the corresponding
quasi-safe images.

Incomplete verifiers are still vulnerable if we allow arbitrary network
architectures and weights. Our exploitation builds on the observation that
verifiers typically need to merge always-active ReLU units with their subsequent
layers to reduce the number of nonlinearities and achieve a reasonable speed.
The merge of layers involves computing merged ``equivalent'' weights, which is
different from the floating point computation adopted by an inference
implementation.

We build a neural network that takes a $13\times 13$ single-channel input image,
followed by a $5\times 5$ convolutional layer with a single output channel, two
fully connected layers with $16$ output neurons each, a fully connected layer
with one output neuron denoted as $u = \max(\V{W_u}\V{h_u} + b_u,\, 0)$, and a
final linear layer that computes $\vy = [u;\, 10^{-7}]$ as the logits vector.
All the hidden layers have ReLU activation. The input $\vxz$ is taken from a
Gaussian distribution. The hidden layers have random Gaussian coefficients, and
the biases are chosen so that
\begin{enuminline}
    \item the ReLU neurons before $u$ are always activated for inputs in the
        perturbation space of $\vxz$
    \item the neuron $u$ is never activated while $b_u$ is the maximum possible
        value (i.e., $b_u = -\max_{\vx\in\adv{x_0}}\V{W_u}\V{h_u}(\vx)$)
\end{enuminline}.
\verb|CROWN| is able to prove that all ReLU neurons before $u$ are always
activated but $u$ is never activated, and therefore it claims that the network
is robust with respect to perturbations around $\vxz$. However, by initializing
the quasi-adversarial input $\vxo \gets \vxz + \epsilon \operatorname{sign}(
\V{W_{equiv}})$ where $\V{W_{equiv}}$ is the product of all the coefficient
matrices of the layers up to $u$, we successfully find adversarial inputs for
all the five implementations considered in this work by randomly perturbing
$\vxo$ using \cref{alg:rnd-pert-adv} with a larger number of iterations
($N=10000$) due to the smaller input size.

Note that the output scores can be manipulated to appear less suspicious. For
instance, we can set $\V{z} = \operatorname{clip}(10^7 \cdot \vy,\, -2
,\,2)$ as the final output in the above example so that $\V{z}$ becomes a more
``naturally looking'' classification score in the range $[-2,\,2]$ and its
perturbation due to floating point error is also enlarged to the unit scale. The
extreme constants $10^{-7}$ and $10^7$ can also be obfuscated by using multiple
consecutive scaling layers with each one having small scaling factors such as
$0.1$ and $10$.


\section{Discussion}

We have shown that some neural network verifiers are systematically exploitable.
One appealing remedy is to introduce relaxations into complete verifiers, such
as by verifying for a larger $\epsilon$ or setting a threshold for accepted
confidence score. For example, it might be tempting to claim the robustness of a
network for $\epsilon=0.09999$ if it is verified for $\epsilon=0.1$. We
emphasize that there are no guarantees provided by any floating point complete
verifier currently extant. Moreover, the difference between the true robust
perturbation bound and the bound claimed by an unsound verifier might be much
larger if the network has certain properties.  For example, \verb|MIPVerify| has
been observed to give NaN results when verifying pruned neural networks~\citep{
guidotti2020verification}.  The adversary might also be able to manipulate the
network to scale the scores arbitrarily, as discussed in
\cref{sec:exploit-incomplete}. The correct solution requires obtaining a tight
relaxation bound that is sound for both the verifier and the inference
implementation, which is extremely challenging.

A possible fix for complete verification is to adopt exact MILP solvers with
rational inputs \citep{steffy2013valid}. There are three challenges:
\begin{enuminline}
    \item The efficiency of exactly solving the large amounts of computation in
        neural network inference has not been studied and is unlikely to be
        satisfactory
    \item The computation that derives the MILP formulation from a verification
        specification, such as the neuron bound analysis in \citet{
        tjeng2018evaluating}, must also be exact, but existing neural network
        verifiers have not attempted to define and implement exact arithmetic
        with the floating point weights
    \item The results of exact MILP solvers are only valid for an exact neural
        network inference implementation, but such exact implementations are not
        widely available (not provided by any deep learning libraries that we
        are aware of), and their efficiency remains to be studied
\end{enuminline}.

Alternatively, one may obtain sound and nearly complete verification by adopting
a conservative MILP solver based on techniques such as directed rounding \citep{
neumaier2004safe}. We also need to ensure all arithmetic in the verifier to
derive the MILP formulation soundly over-approximates floating point error.
This is more computationally feasible than exact verification discussed above.
It is similar to the approach used in some sound incomplete verifiers that
incorporate floating point error by maintaining upper and lower rounding bounds
of internal computations~\citep{ singh2018fast, singh2019an}. However, this
approach relies on the specific implementation details of the inference
algorithm --- optimizations such as Winograd~\citep{ lavin2016fast} or
FFT~\citep{ abtahi2018accelerating}, or deployment in hardware accelerators with
lower floating point precision such as Bfloat16 \citep{burgess2019bfloat16},
would either invalidate the robustness guarantees or require changes to the
analysis algorithm. Therefore, we suggest that these sound verifiers explicitly
state the requirements on the inference implementations for which their results
are sound.  A possible future research direction is to devise a universal sound
verification framework that can incorporate different inference implementations.

Another approach for sound and complete neural network verification is to
quantize the computation to align the inference implementation with the
verifier. For example, if we require all activations to be multiples of $s_0$
and all weights to be multiples of $s_1$, where $s_0s_1 > 2E$ and $E$ is a very
loose bound of possible implementation error, then the output can be rounded to
multiples of $s_0s_1$ to completely eliminate numerical error. Binarized neural
networks~\citep{hubara2016binarized} are a family of extremely quantized
networks, and their verification~\citep{ narodytska2018verifying,
shih2019verifying, jia2020efficient} is sound and complete. However, the problem
of robust training and verification of quantized neural networks~\citep{
jia2020efficient} is relatively under-examined compared to that of real-valued
neural networks~\citep{ madry2018towards, mirman2018differentiable,
tjeng2018evaluating, xiao2018training}.


\section{Conclusion}

Floating point error should not be overlooked in the verification of real-valued
neural networks, as we have presented techniques that efficiently find witnesses
for the unsoundness of two verifiers. Unfortunately, floating point soundness
issues have not received sufficient attention in neural network verification
research. A user has few choices if they want to obtain sound verification
results for a neural network, especially if they deploy accelerated neural
network inference implementations. We hope our results will help to guide future
neural network verification research by providing another perspective on the
tradeoff between soundness, completeness, and scalability.


\subsubsection*{Acknowledgments}
We would like to thank Gagandeep Singh and Kai Xiao for providing invaluable
suggestions on an early manuscript.



%
%
\bibliographystyle{splncs04nat}
{
    \small
    \bibliography{references}
}
\end{document}